\documentclass{article}
\PassOptionsToPackage{numbers, compress}{natbib}

\usepackage{graphicx}
\usepackage{longtable}
 \usepackage[preprint]{neurips_2026}

\usepackage[utf8]{inputenc} 
\usepackage[T1]{fontenc}    
\usepackage{hyperref}       
\usepackage{url}            
\usepackage{booktabs}       
\usepackage{amsfonts}       
\usepackage{nicefrac}       
\usepackage{microtype}      
\usepackage{xcolor}         

\title{Key-Gram: Extensible World Knowledge for Embodied Manipulation}

\author{%
  Jingjing Fan\textsuperscript{1,}\thanks{Equal contribution.}, 
  Siyuan Li\textsuperscript{2, $\ast$}, 
  Botao Ren\textsuperscript{1}, \textnormal{and} 
  Zhidong Deng\textsuperscript{1,}\thanks{Corresponding author.}\\
  \textsuperscript{1}Department of Computer Science and Technology \quad
  \textsuperscript{2}Department of Automation \\
  Tsinghua University \\
  \texttt{fan-jj24@mails.tsinghua.edu.cn} 
}

\begin{document}

\maketitle

\begin{abstract}
Embodied control increasingly requires models to follow compositional language instructions while reasoning over dynamic visual states. However, current vision-language-action policies and world-action models often couple linguistic knowledge with visual computation in a shared backbone or conditioning pathway, leading to modality competition and making knowledge extension dependent on backbone updates. In this paper, we introduce Key-Gram, a conditional-memory framework that separates language-derived world knowledge from visual-state reasoning for embodied control. At its core is a memory module that decomposes an instruction into task-specific key-grams, retrieves static linguistic priors through deterministic hashed lookup, and injects the retrieved entries into selected hidden layers through context-aware gating and lightweight convolutional fusion. This design allows the backbone to devote its main capacity to visual reasoning and action inference, while reusable instruction knowledge is stored in an extensible external memory. The logical memory table can be conveniently partitioned during training and, due to its $O(1)$ lookup pattern, efficiently placed on host memory during inference. Across RoboTwin2.0, LIBERO/LIBERO-Plus, and real-world dual-arm manipulation, Key-Gram consistently improves both $\pi_{0}$ and $\pi_{0.5}$ backbones, with average relative gains of $29.5\%/9.9\%$ on RoboTwin2.0, $35.8\%/4.5\%$ on LIBERO-Plus transfer without target-domain fine-tuning, and $15.4\%/8.1\%$ on real-world long-horizon tasks. These results demonstrate that externalized linguistic memory provides an effective and extensible mechanism for improving compositional grounding, transfer, and real-world manipulation.
\end{abstract}

\section{Introduction}

Embodied manipulation places two fundamentally different demands on a single model: it must preserve reusable world knowledge about objects, skills, relations, constraints, and task structure, while continuously adapting its internal model of how the visual world evolves under action. Recent vision-language-action (VLA) models have made instruction-sensitive control practical by bringing language priors into end-to-end policies \cite{zitkovich2023rt2, kim2025openvla, black2024pi0, black2025pi05, kim2025openvlaoft, liu2024rdt, bjorck2025groot, cheang2025gr3, zheng2025xvla}, while recent World Action Models (WAM) have highlighted the value of forecasting future world states for physical generalization \cite{cheang2024gr2, huang2025enerverse, liao2025genie, lu2025gwm, li2026lingbotva, ye2026dreamzero}. Yet both trends sharpen the same unresolved tension: static world knowledge and online physical reasoning remain tightly entangled inside the policy computation itself.

Existing efforts largely approach this tension in two ways. The first is dense fusion, dominant in mainstream VLA formulations, where language and visual observations are projected into a shared token space and processed by the same backbone \cite{zitkovich2023rt2, kim2025openvla, black2024pi0}. This design is simple and scalable, but it forces a small number of instruction tokens to compete with a much larger and continuously changing set of visual tokens inside the same dense attention stream. The second is conditional generative modeling, increasingly adopted in diffusion-based robot policies and WAMs, where language is encoded separately and injected as conditioning to guide the generation of future visual states, actions, or both \cite{chi2023diffusionpolicy, ye2026dreamzero}. Community efforts such as FiLM-style modulation, cross-attention, and AdaLN-based conditioning should be understood as local refinements within this broader line rather than as a distinct paradigm of their own \cite{perez2018film, dumoulin2018featurewise, peebles2023dit, dasari2024ingredients}: they acknowledge that naive fusion is problematic, but they still leave unresolved a deeper question---what role language should play in embodied control. In dense fusion, language is drawn into token-level competition at the input interface; in conditional generative models, it is reduced to a prompt-like steering signal, while compositional world knowledge remains implicitly baked into the generative backbone. In both cases, the architecture still fails to separate reusable knowledge from online scene reasoning, but merely entangles them at different locations. Consequently, when such models adapt to new physical regimes, the gradient updates that improve online reasoning can also overwrite previously acquired world knowledge, making continual adaptation and modular extension inherently fragile \cite{mccloskey1989catastrophic, french1999catastrophic}.

We take a different view. In embodied control, visual computation should primarily reason over scene dynamics: what state the world is currently in, what future should be reached, and how action should evolve accordingly. Language, by contrast, does not primarily specify low-level physical evolution; rather, it serves as a compact and reusable index over abstract task priors. This functional asymmetry suggests that embodied models should separate information by function rather than by modality. Motivated by conditional-memory formulations \cite{cheng2026engram}, we introduce Key-Gram, in which the instruction is decomposed into a small set of task-specific key-grams that retrieve an external linguistic memory. The retrieved memory are then injected into the backbone as reusable priors, while its dominant computation remains focused on future-state reasoning and control-relevant physical evolution. More importantly, this design induces a principle of extensibility. When new knowledge is acquired, it can be appended as new entries in the external memory rather than rewritten into backbone weights. The backbone may still update to learn how newly retrieved knowledge should be grounded and used in physical interaction. What no longer needs to change, however, is the previously acquired world knowledge itself. In this way, knowledge growth becomes modular, while existing memory is protected from gradient interference during backbone adaptation, making the architecture naturally compatible with open-world deployment and continual accumulation.

Empirically, this functional decoupling brings consistent improvements across simulated and real-world manipulation. Across RoboTwin2.0 \cite{chen2025robotwin2}, LIBERO \cite{liu2023libero}, LIBERO-Plus \cite{fei2025liberoplus}, and real-world manipulation, Key-Gram improves both $\pi_{0}$ and $\pi_{0.5}$ backbones, with average relative gains of $29.5\%/9.9\%$ on RoboTwin2.0, $35.8\%/4.5\%$ on LIBERO-Plus transfer without target-domain fine-tuning, and $15.4\%/8.1\%$ on real-world long-horizon tasks. The gains are especially strong in instruction-sensitive settings where similar visual scenes require different linguistic grounding. In real-world expansion tasks, Key-Gram substantially improves unseen compositional pairings, increasing Task~3 by $34.6\%/18.8\%$ and Task~4 by $41.7\%/21.2\%$, and also improves sequential adaptation in Task~5 by $10.0\%/4.7\%$. These results suggest that separating language-side world-knowledge retrieval from vision-side physical reasoning provides a more effective inductive bias for embodied intelligence than either dense token fusion or coarse generative conditioning.

Our contributions are summarized as follows:
\begin{itemize}
    \item \textbf{A functionally decoupled embodied framework.}
    We propose Key-Gram, which separates instruction-side world-knowledge retrieval from vision-side physical reasoning, reducing modality competition in embodied control.

    \item \textbf{An extensible external memory for world knowledge.}
    Key-Gram stores reusable linguistic priors in a structured memory accessed by deterministic hashed lookup, enabling modular knowledge expansion without rewriting the backbone.

    \item \textbf{Consistent gains in instruction-sensitive manipulation.}
    Experiments across simulated and real-world benchmarks show improved compositional grounding, transfer, and adaptation under diverse linguistic formulations.
\end{itemize}

\begin{figure}[t]
  \centering
  \includegraphics[width=\linewidth]{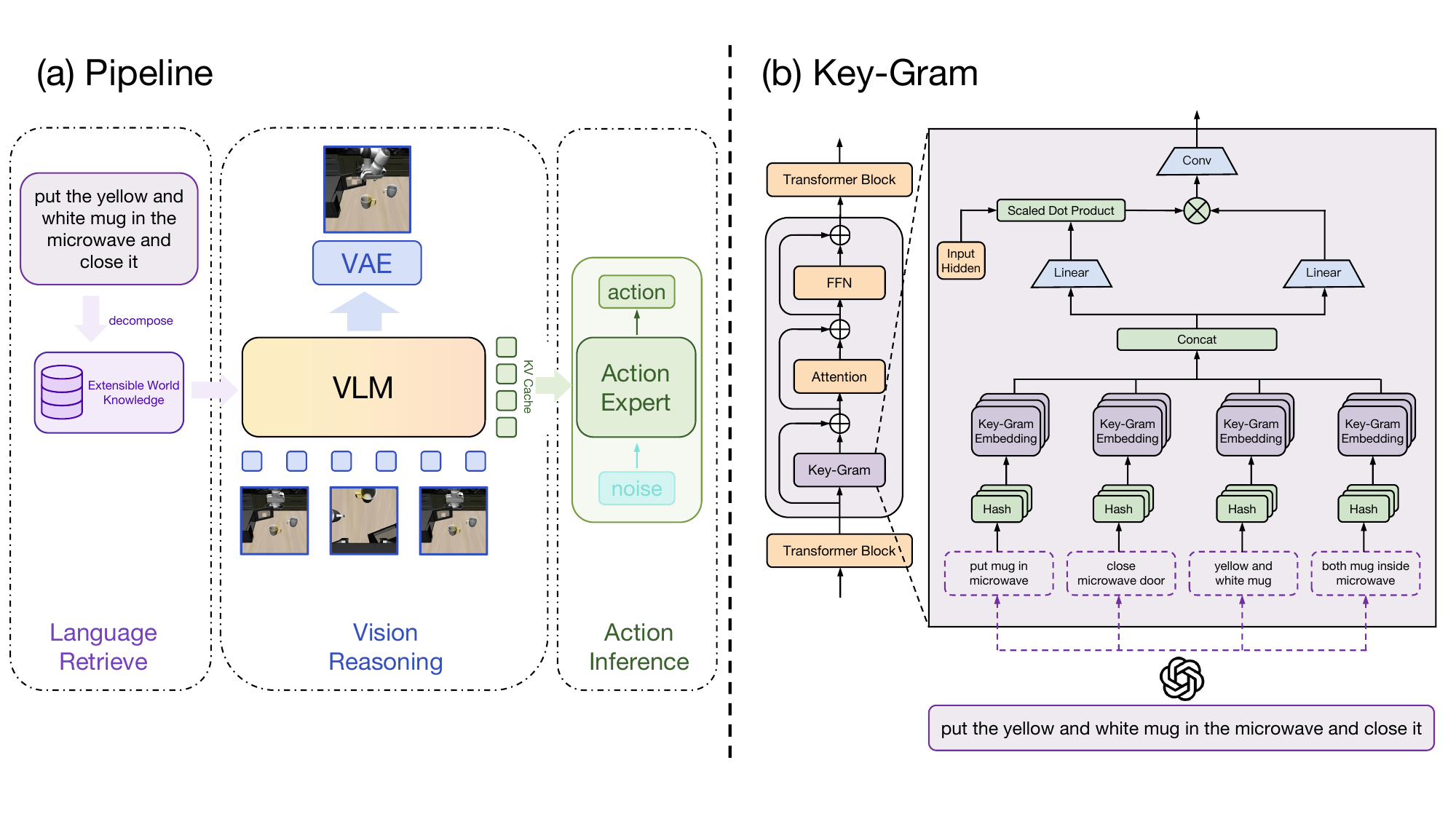}
  \caption{
  Overview of Key-Gram. 
  (a) The framework separates language-derived knowledge retrieval, visual reasoning, and action inference. 
  (b) The Key-Gram module retrieves external linguistic priors from decomposed key-grams via multi-head hashing, and injects them into selected Transformer layers through context-aware gated fusion.
  }
  \label{fig:keygram-overview}
\end{figure}

\section{Related Work}

\subsection{Vision-Language-Action Models}

VLA models have become a major paradigm for language-conditioned robot control by transferring large-scale vision-language priors into visuomotor policies. Systems such as RT-2, OpenVLA, $\pi_{0}$, $\pi_{0.5}$, and X-VLA show that scaling vision-language and robot data improves semantic grounding, instruction following, cross-embodiment transfer, and long-horizon manipulation \cite{zitkovich2023rt2, kim2025openvla, black2024pi0, black2025pi05, zheng2025xvla}.

Despite these advances, most VLA models adopt dense multimodal fusion: language, visual, and sometimes action tokens are embedded into a shared token space and processed by a common backbone \cite{zitkovich2023rt2, kim2025openvla, black2024pi0}. This simple recipe creates a structural bottleneck for manipulation, as short instructions must compete with dense visual tokens within the same attention stream. Such modality competition can dilute instruction-specific information in cluttered or compositional scenes. In contrast, Key-Gram separates instruction-side world-knowledge retrieval from scene-side visual reasoning, thereby bypassing input-level token competition.

\subsection{World Action Models and Predictive Manipulation}

A parallel line of work argues that robust manipulation requires not only semantic grounding, but also explicit prediction of future observations, latent world states, or action trajectories. This view motivates diffusion-based visuomotor policies and the recent WAM paradigm \cite{chi2023diffusionpolicy, lu2025gwm, li2026lingbotva, ye2026dreamzero}. For example, DreamZero adapts a large pretrained video diffusion model into a real-time closed-loop controller, showing that future-state prediction can provide a strong inductive bias for long-horizon manipulation \cite{ye2026dreamzero}.

However, WAM-style methods mainly strengthen dynamic scene prediction, while leaving world knowledge implicit in the same monolithic model. In embodied settings, such knowledge is open-ended, as new objects, skills, and relations continually emerge. Once static knowledge and dynamic prediction are entangled in a single backbone, extension becomes costly and brittle. Key-Gram addresses this limitation by decoupling expandable world knowledge from visual future-state reasoning.

\subsection{World Memory, Layer-wise Memory Scaling, and Lifelong Expansion}

Recent LLM research increasingly treats world knowledge as an explicit memory primitive rather than as information fully absorbed by dense backbone computation. Classical memory-augmented models, such as product-key memories and retrieval-augmented language models, decouple storage from computation for greater capacity and efficiency \cite{lample2019pkm, guu2020realm, lewis2020rag, borgeaud2022retro}. Recent work further develops this direction through hashed conditional memory, large static embedding tables, layer-wise memory modules, and per-layer embedding tables \cite{wu2022memorizing, wang2023longmem, cheng2026engram, liu2026scalingembeddings, ding2026meki, gemma42026docs}.

Yet these methods are primarily designed for language modeling, scaling, or efficient inference, not for embodied lifelong learning. In manipulation, the key issue is continual knowledge expansion: new objects, skills, and relations should be added without repeatedly rewriting the policy backbone. When such knowledge is entangled with core parameters, adaptation becomes expensive and prone to catastrophic forgetting \cite{mccloskey1989catastrophic, french1999catastrophic}. Key-Gram addresses this limitation by organizing world knowledge as an incrementally expandable external memory, enabling modular knowledge growth for embodied control without full-model retraining.

\section{Method}

\subsection{Overview}
\label{sec:method-overview}

We introduce Key-Gram, an external conditional memory module that augments visual reasoning models with structured linguistic world knowledge. Given a language instruction $\mathcal{I}$ and initial visual observations $\mathcal{V}_{0}$, the model predicts both an action trajectory and a compact future visual state. As shown in Fig.~\ref{fig:keygram-overview}, Key-Gram separates instruction-side knowledge retrieval from scene-side visual reasoning. It first decomposes $\mathcal{I}$ into a small set of task-specific key-grams and maps them to static dense memory embeddings through deterministic lookup. The retrieved embeddings are then injected into selected backbone layers via context-adaptive modulation and lightweight convolutional refinement.

The memory-guided backbone supports two downstream prediction pathways: final-layer hidden states are adapted into VAE-encodable future latents, while the backbone KV cache conditions an action expert for trajectory decoding. This design keeps the main visual backbone focused on future-state reasoning and control, while world knowledge is stored in an expandable external memory. The following sections describe key-gram retrieval, memory fusion, prediction heads, extensibility, and the training--inference system design.
 
\subsection{Sparse Key-Gram Retrieval}
\label{sec:keygram-retrieval}

The first phase maps instruction-level semantic units to static memory entries through key-gram extraction and deterministic hashed retrieval.

\paragraph{Key-gram extraction.}
Rather than relying on tokenization or contiguous phrase segmentation, we use a lightweight language model or API-based parser to extract a fixed number of short, knowledge-bearing key-grams from each instruction. The parser is prompted to compose reusable task-level units under a maximum word length. For example, ``put the yellow and white mug in the microwave and close it'' can be decomposed into \{\textit{put mug in microwave}, \textit{close microwave door}, \textit{yellow and white mug}, \textit{mug inside microwave}\}. The key-gram budget is set according to the instruction complexity of each environment, and the extraction prompt is provided in Appendix~\ref{app:keygram-prompt}.
\paragraph{Hashed memory mapping.}
Let $\mathcal{G}=\{g_i\}_{i=1}^{K}$ be the extracted key-grams, where each $g_i$ contains at most $M$ words. We convert each word into an integer identifier and right-pad shorter key-grams with zeros to obtain a fixed-length key $\bar{g}_i=(a_{i,1},\ldots,a_{i,M})$. Following multiplicative-XOR hashing, each key is mapped to $H$ hash heads. For the $h$-th head at layer $\ell$, the memory index is computed as
\begin{equation}
z^{(\ell,h)}_i
=
\left(
\bigoplus_{j=1}^{M}
a_{i,j} r^{(\ell,h)}_{j}
\right)
\bmod P^{(\ell,h)},
\end{equation}
where $\oplus$ denotes bitwise XOR, $r^{(\ell,h)}_{j}$ are deterministic odd multipliers, and $P^{(\ell,h)}$ is a prime table size. The corresponding embedding is retrieved as $e^{(\ell,h)}_i=E^{(\ell,h)}[z^{(\ell,h)}_i]$, and all head outputs are concatenated to form
\begin{equation}
e^{(\ell)}_i=\mathop{\Vert}_{h=1}^{H} e^{(\ell,h)}_i .
\end{equation}
The retrieved embeddings $\{e^{(\ell)}_i\}_{i=1}^{K}$ are used as static linguistic memory for the subsequent fusion stage.

\subsection{Context-Adaptive Key-Gram Fusion}
\label{sec:keygram-fusion}

As illustrated in Fig.~\ref{fig:keygram-overview} (b), the fusion phase injects retrieved key-gram memories into the visual backbone through a residual module inserted before the attention operation of selected Transformer blocks.

Let $H^{(\ell)}\in\mathbb{R}^{B\times L\times d}$ denote the hidden states before attention at layer $\ell$, where $L$ is the number of visual tokens. The retrieved embeddings from Sec.~\ref{sec:keygram-retrieval} are first concatenated into a memory vector $M^{(\ell)}\in\mathbb{R}^{B\times 1\times d_m}$, where $d_m$ aggregates both the $K$ extracted key-grams and their $H$ hash heads. Specifically, if each hash head returns a $d_h$-dimensional vector, then $d_m=K H d_h$. We therefore project the memory into the backbone space through two learnable matrices, yielding $K_m^{(\ell)}=M^{(\ell)}W_K^{(\ell)}$ and $V_m^{(\ell)}=M^{(\ell)}W_V^{(\ell)}$, where $K_m^{(\ell)},V_m^{(\ell)}\in\mathbb{R}^{B\times 1\times d}$. The incoming hidden states are directly used as visual queries. A token-wise gate is computed as
\begin{equation}
A^{(\ell)}
=
\sigma\!\left(
H^{(\ell)} {K_m^{(\ell)}}^\top / \sqrt{d}
\right),
\qquad
A^{(\ell)}\in\mathbb{R}^{B\times L\times 1}.
\end{equation}
This gate measures the relevance of the retrieved linguistic memory to each visual token, enabling the model to inject knowledge priors only where they are useful.

The gated memory feature is then refined by a lightweight long-span convolution:
\begin{equation}
\Delta H^{(\ell)}
=
\mathrm{Conv}_{\mathrm{span}}
\!\left(
A^{(\ell)}\odot V_m^{(\ell)}
\right).
\end{equation}
Since $M^{(\ell)}$ is formed by concatenating multiple key-gram memories, the projected value $V_m^{(\ell)}$ contains heterogeneous pieces of instruction-level knowledge. The long-span convolution is used to mix these memory components over a wider receptive field, allowing relations among different key-grams to be jointly modeled before being injected into each visual token. This operation provides an explicit interaction mechanism among retrieved knowledge entries beyond independent gating.

The Key-Gram output is finally added to the original hidden states in residual form, $\widetilde{H}^{(\ell)}=H^{(\ell)}+\Delta H^{(\ell)}$, and $\widetilde{H}^{(\ell)}$ is passed to the standard attention computation. We instantiate this module only at selected backbone layers, following the placement strategy validated by the ablation study in Sec.~\ref{sec:exp-ablation}.

\begin{figure*}[t]
    \centering
    \includegraphics[width=\linewidth]{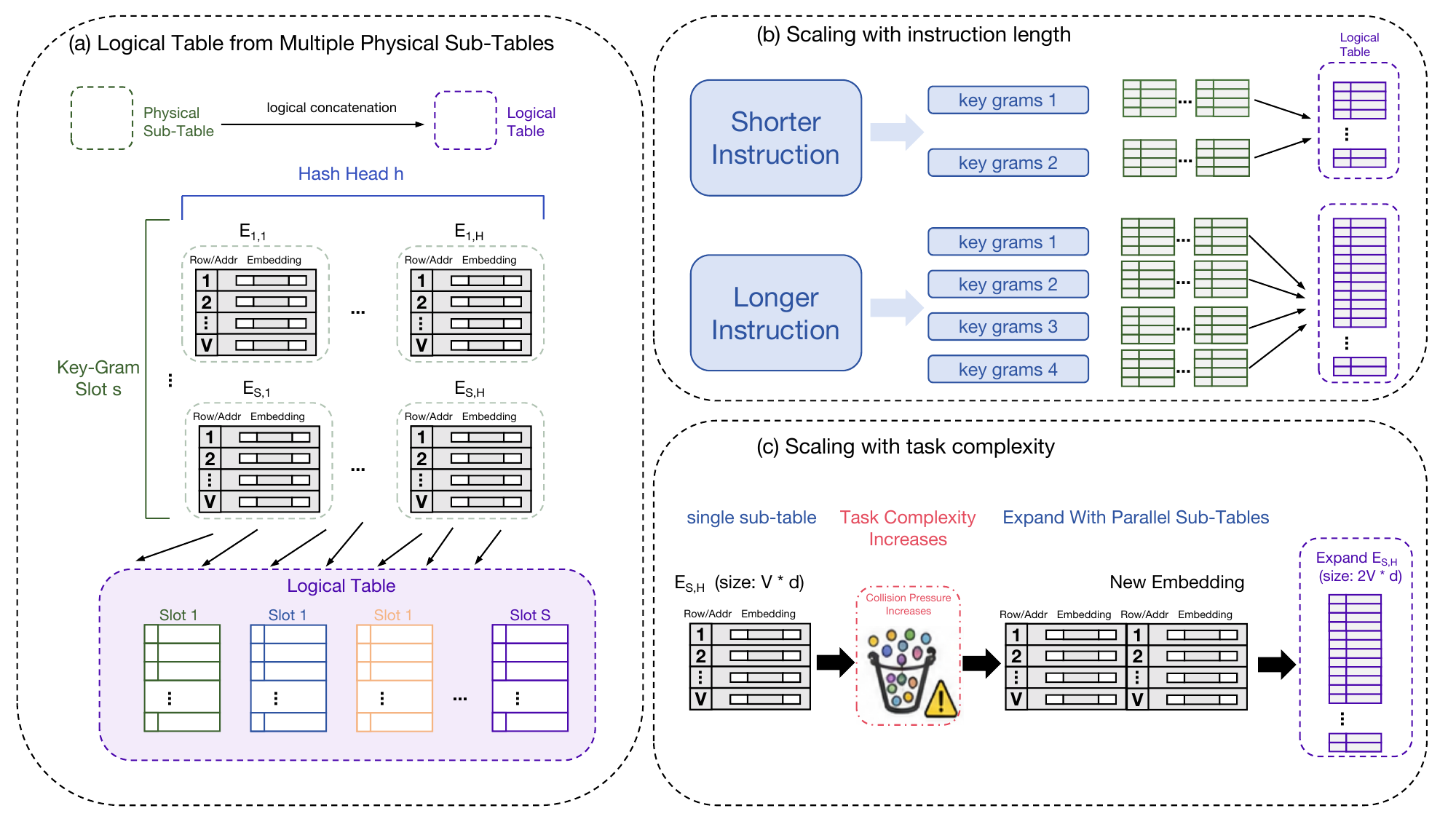}
    \caption{
    Extensible memory allocation of Key-Gram.
    The memory is a logical table composed of slot-head physical sub-tables rather than a single monolithic table.
    Longer instructions are handled by adding key-gram slots, while more complex task distributions are supported by appending new sub-tables to enlarge the effective capacity and reduce hash collisions.
    }
    \label{fig:keygram-extensibility}
\end{figure*}

\subsection{Coupled Visual and Action Prediction}
\label{sec:keygram-prediction}

As illustrated in Fig.~\ref{fig:keygram-overview} (a), the memory-guided backbone is followed by two prediction pathways for visual foresight and action decoding.

\paragraph{Vision reasoning.}
Let $H^{(L)}$ denote the final-layer hidden states of the backbone. We introduce a short sequence of learnable queries $Q_v$ and use a cross-attention block to extract compact future-vision representations from $H^{(L)}$. The resulting query features are projected into the visual latent space and passed through two lightweight upsampling stages to match the spatial resolution of the VAE latent. The visual pathway therefore predicts visual foresight in a compact latent space.

\paragraph{Action inference.}
For action prediction, we reuse the KV cache accumulated by the preceding backbone as conditional context for a lightweight action expert. The action expert is architecturally isomorphic to the backbone but smaller in scale, and performs a flow-matching denoising process over action trajectories. This design allows action decoding to remain tightly coupled with the same memory-guided visual reasoning process that produces future visual latents.

\subsection{Extensible and System-Efficient Memory Allocation}
\label{sec:keygram-extensibility}

As illustrated in Fig.~\ref{fig:keygram-extensibility}, Key-Gram organizes memory as a logical table composed of multiple sub-tables rather than a single monolithic table. Each key-gram slot and hash head owns an independent sub-table $E_{s,h}\in\mathbb{R}^{V\times d_e}$, where $V$ is the row capacity and $d_e$ is the embedding width. Retrieved entries from all slot-head sub-tables are concatenated before fusion. This structure supports both flexible memory expansion and efficient system realization: during training, sub-tables can be partitioned across devices under the same deterministic lookup interface; during inference, indices are determined solely by the instruction and fixed hash functions, enabling $O(1)$ lookup with respect to total memory size and allowing large knowledge tables to reside in CPU memory.

\paragraph{Scaling with instruction length.}
Longer instructions typically require more key-grams to represent their world knowledge. Key-Gram handles this by increasing the number of key-gram slots, where each added slot introduces a group of $H$ hash-head sub-tables. For example, a short instruction may activate key-grams such as \{\textit{put mug in microwave}, \textit{close microwave door}, \textit{yellow and white mug}\}, whereas a longer instruction may require additional entries such as \{\textit{pick hamburg and fries}, \textit{French fries in red packaging}, \textit{smooth glossy orange tray}\}. Thus, longer tasks can retrieve richer linguistic priors while keeping the visual backbone unchanged.

\paragraph{Scaling with task complexity.}
More complex environments introduce additional entities, relations, and affordances, increasing hash-collision pressure under a fixed memory capacity. Key-Gram expands capacity by enlarging $V$ or adding parallel sub-tables to existing slot-head groups, whose outputs are concatenated with the original retrieved entries. This increases the addressable knowledge space without modifying the visual backbone, allowing new knowledge to be appended while preserving existing memory structure and backbone parameters.

\begin{table*}[t]
  \caption{RoboTwin2.0 results (\%). Gains in parentheses for Key-Gram variants are relative improvements over their corresponding base backbones.}
  \label{tab:robotwin-results}
  \centering
  \small
  \setlength{\tabcolsep}{3pt}
  \begin{tabular}{lcccccccccc}
    \toprule
    & \multicolumn{2}{c}{X-VLA}
    & \multicolumn{2}{c}{$\pi_{0}$}
    & \multicolumn{2}{c}{$\pi_{0.5}$}
    & \multicolumn{2}{c}{$\pi_{0}$-KG (ours)}
    & \multicolumn{2}{c}{$\pi_{0.5}$-KG (ours)} \\
    \cmidrule(r){2-3}
    \cmidrule(r){4-5}
    \cmidrule(r){6-7}
    \cmidrule(r){8-9}
    \cmidrule(r){10-11}
    Metric
      & Easy & Hard
      & Easy & Hard
      & Easy & Hard
      & Easy & Hard
      & Easy & Hard \\
    \midrule
    Avg. $_{\mathrm{H}=1}$
      & 81.6 & 82.5
      & 66.5 & 61.6
      & 85.1 & 80.2
      & 80.4 {\scriptsize(+20.9\%)}
      & 78.0 {\scriptsize(+26.4\%)}
      & 90.8 {\scriptsize(+6.6\%)}
      & 87.0 {\scriptsize(+8.5\%)} \\
    Avg. $_{\mathrm{H}=2}$
      & 59.3 & 55.9
      & 66.1 & 54.7
      & 79.3 & 73.0
      & 80.5 {\scriptsize(+21.7\%)}
      & 72.7 {\scriptsize(+33.0\%)}
      & 86.3 {\scriptsize(+8.7\%)}
      & 80.3 {\scriptsize(+10.0\%)} \\
    Avg. $_{\mathrm{H}=3}$
      & 61.2 & 66.0
      & 61.6 & 50.2
      & 78.6 & 67.4
      & 79.4 {\scriptsize(+28.9\%)}
      & 70.2 {\scriptsize(+39.8\%)}
      & 86.8 {\scriptsize(+10.4\%)}
      & 80.8 {\scriptsize(+19.9\%)} \\
    Avg. $_{\mathrm{50\ Tasks}}$
      & 72.9 & 72.8
      & 65.9 & 58.4
      & 82.7 & 76.8
      & 80.3 {\scriptsize(+21.9\%)}
      & 75.6 {\scriptsize(+29.5\%)}
      & 89.0 {\scriptsize(+7.6\%)}
      & 84.4 {\scriptsize(+9.9\%)} \\
    \bottomrule
  \end{tabular}
\end{table*}

\begin{table*}[t]
  \caption{LIBERO and LIBERO-Plus results (\%). Gains in parentheses for Key-Gram variants are relative improvements over their corresponding base backbones.}
  \label{tab:libero-results}
  \centering
  \small
  \setlength{\tabcolsep}{4pt}
  \begin{tabular}{lccc}
    \toprule
    Model
      & LIBERO fine-tuned
      & LIBERO-Plus from LIBERO
      & LIBERO-Plus fine-tuned \\
    \midrule
    OpenVLA-OFT
      & 95.3 & 69.6 & 79.6 \\
    $\pi_{0}$
      & 94.2 & 53.6 & 84.0 \\
    $\pi_{0.5}$
      & 96.9 & 83.9 & 90.4 \\
    \midrule
    $\pi_{0}$-KG (ours)
      & 94.6 {\scriptsize(+0.4\%)}
      & 72.8 {\scriptsize(+35.8\%)}
      & 88.5 {\scriptsize(+5.4\%)} \\
    $\pi_{0.5}$-KG (ours)
      & 96.7 {\scriptsize(-0.2\%)}
      & 87.7 {\scriptsize(+4.5\%)}
      & 92.6 {\scriptsize(+2.4\%)} \\
    \bottomrule
  \end{tabular}
\end{table*}

\section{Experiment}
\subsection{Experiments Setup}

\paragraph{Benchmarks.}
We evaluate $\pi_{0}$-Key-Gram and $\pi_{0.5}$-Key-Gram on RoboTwin2.0, LIBERO, and LIBERO-Plus \cite{chen2025robotwin2, liu2023libero, fei2025liberoplus}. Following the RoboTwin2.0 protocol, we test all 50 tasks under both clean and randomized settings, yielding 100 configurations with 100 trials each in unseen environments. For LIBERO, we report results on the four standard task suites. For LIBERO-Plus, we follow its distribution-expansion and robustness setting, evaluating both zero-shot transfer from LIBERO-trained models and further fine-tuned models, measuring in-domain performance and generalization to expanded task variations.

\paragraph{Model Configuration.}
We instantiate Key-Gram on two open-source VLA backbones, $\pi_{0}$ and $\pi_{0.5}$ \cite{black2024pi0, black2025pi05}, while keeping the original backbone and action expert unchanged. Key-Gram is inserted as a lightweight residual module into Transformer Layers $(1,8,13)$, following the layer-placement ablation in Sec.~\ref{sec:exp-ablation}. This setting isolates the effect of external linguistic memory from changes to the base VLA architecture. Detailed memory size, hashing, and fusion configurations are provided in Appendix~\ref{appendix:model-config}.

\paragraph{Real-world robot.}
We conduct real-world experiments on a Piper dual-arm platform, covering three long-horizon tasks and five expansion-task settings. For long-horizon evaluation, we collect 300 trajectories per task, including plate-based object picking, object sorting into boxes, and compositional object assembly. For expansion evaluation, we collect two 50-trajectory training sets, \emph{pick Pen and Charger} and \emph{pick Matchbox car and Lighter}. We then test in-distribution execution, compositional transfer to unseen pairings \emph{pick Pen and Lighter} and \emph{pick Charger and Matchbox car}, and sequential adaptation by re-evaluating \emph{pick Pen and Charger} after further training on \emph{pick Matchbox car and Lighter}. This protocol evaluates instruction grounding, object-concept recombination, and robustness to knowledge updates.

\begin{table*}[t]
  \caption{Real-world long-horizon task results (\%). Gains in parentheses for Key-Gram variants are relative improvements over their corresponding base backbones.}
  \label{tab:real-long-horizon}
  \centering
  \small
  \setlength{\tabcolsep}{6pt}
  \begin{tabular}{lcccc}
    \toprule
    Model & Picking & Sorting & Assembly & Avg. \\
    \midrule
    $\pi_{0}$
      & 82.0 & 74.0 & 52.0 & 69.3 \\
    $\pi_{0.5}$
      & 90.0 & 86.0 & 70.0 & 82.0 \\
    \midrule
    $\pi_{0}$-KG (ours)
      & 88.0 {\scriptsize(+7.3\%)}
      & 86.0 {\scriptsize(+16.2\%)}
      & 66.0 {\scriptsize(+26.9\%)}
      & 80.0 {\scriptsize(+15.4\%)} \\
    $\pi_{0.5}$-KG (ours)
      & 94.0 {\scriptsize(+4.4\%)}
      & 90.0 {\scriptsize(+4.7\%)}
      & 82.0 {\scriptsize(+17.1\%)}
      & 88.7 {\scriptsize(+8.1\%)} \\
    \bottomrule
  \end{tabular}
\end{table*}

\begin{figure*}[t]
    \centering
    \includegraphics[width=\linewidth]{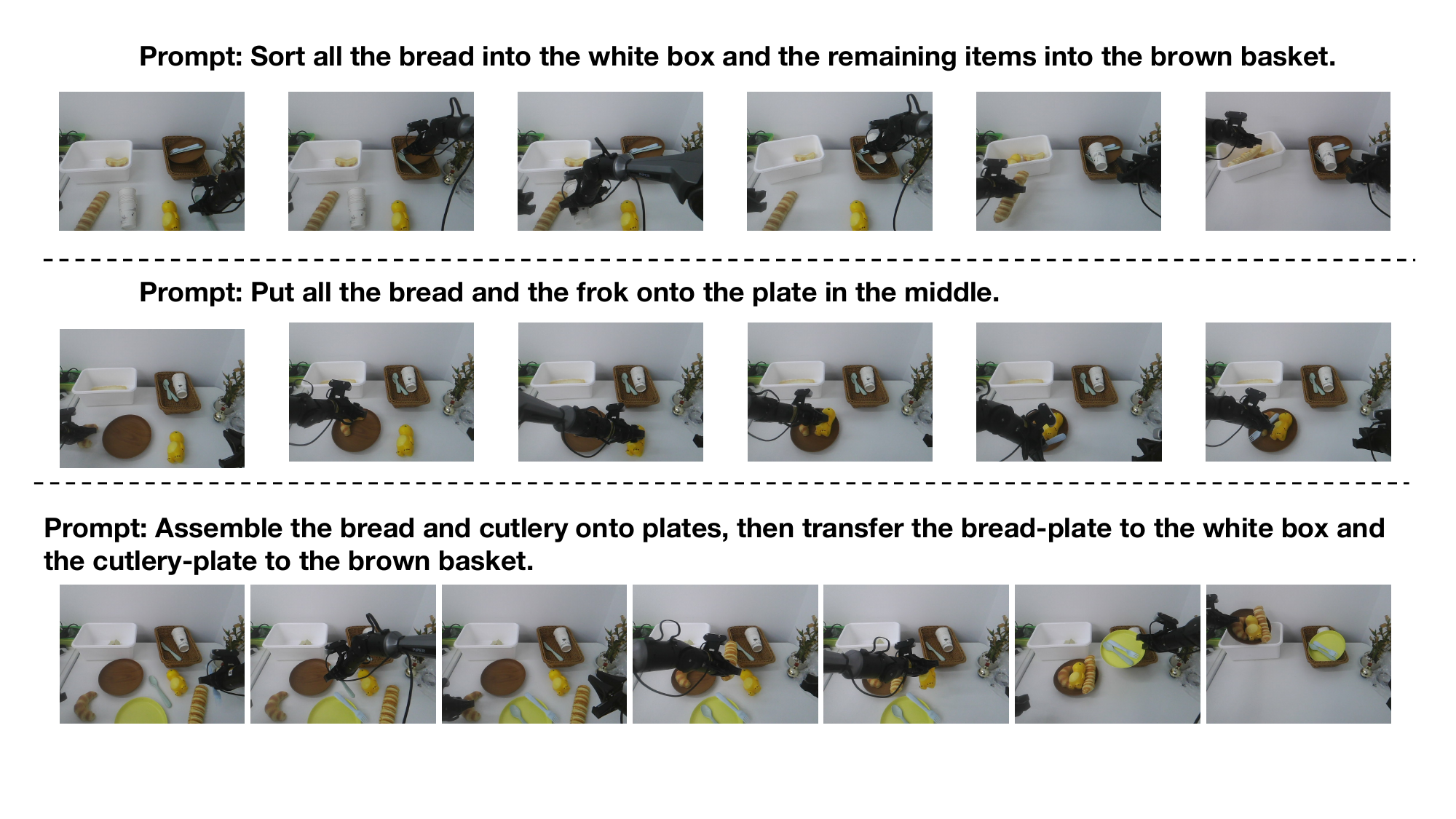}
    \caption{Demonstrations show the execution process of $\pi_{0.5}$-KG.}
    \label{fig:real-long-horizon}
\end{figure*}

\subsection{Results on Simulation}
\label{sec:sim-results}

As shown in Table~\ref{tab:robotwin-results}, Key-Gram consistently improves both $\pi_{0}$ and $\pi_{0.5}$ on RoboTwin2.0. The gains are larger for $\pi_{0}$, where $\pi_{0}$-KG increases the average success rate from 65.9 to 80.3 under the easy setting and from 58.4 to 75.6 under the hard setting. The improvement also grows with task horizon, suggesting that external linguistic memory helps preserve object relations and action priors across multi-step manipulation. On the stronger $\pi_{0.5}$ backbone, Key-Gram still yields stable gains, improving average performance from 82.7 to 89.0 on easy tasks and from 76.8 to 84.4 on hard tasks. Detailed per-task results are provided in Appendix~\ref{app:robotwin-details}.

Table~\ref{tab:libero-results} further shows that Key-Gram improves LIBERO-Plus generalization while preserving strong LIBERO performance. Under direct transfer from LIBERO-trained models, $\pi_{0}$-KG improves over $\pi_{0}$ by 35.8\%, and $\pi_{0.5}$-KG improves over $\pi_{0.5}$ by 4.5\%. After fine-tuning on LIBERO-Plus, both variants continue to outperform their backbones, reaching 88.5 and 92.6, respectively. These results indicate that Key-Gram is especially effective under distribution expansion, where language-derived world priors help bridge original training tasks and more diverse manipulation variations.

\subsection{Results on Real-world Robot}

\paragraph{Long-horizon Tasks.}
As shown in Table~\ref{tab:real-long-horizon}, Key-Gram consistently improves both $\pi_{0}$ and $\pi_{0.5}$ on real-world long-horizon tasks. The largest gains appear on the assembly task, where $\pi_{0}$-KG improves over $\pi_{0}$ by $26.9\%$ and $\pi_{0.5}$-KG improves over $\pi_{0.5}$ by $17.1\%$. This suggests that external linguistic memory is particularly useful for maintaining object relations and sub-goal structure across multiple manipulation stages. Qualitative rollouts in Fig.~\ref{fig:real-long-horizon} show successful execution of sorting, picking, and compositional assembly instructions.

\begin{figure*}[t]
    \centering
    \includegraphics[width=\linewidth]{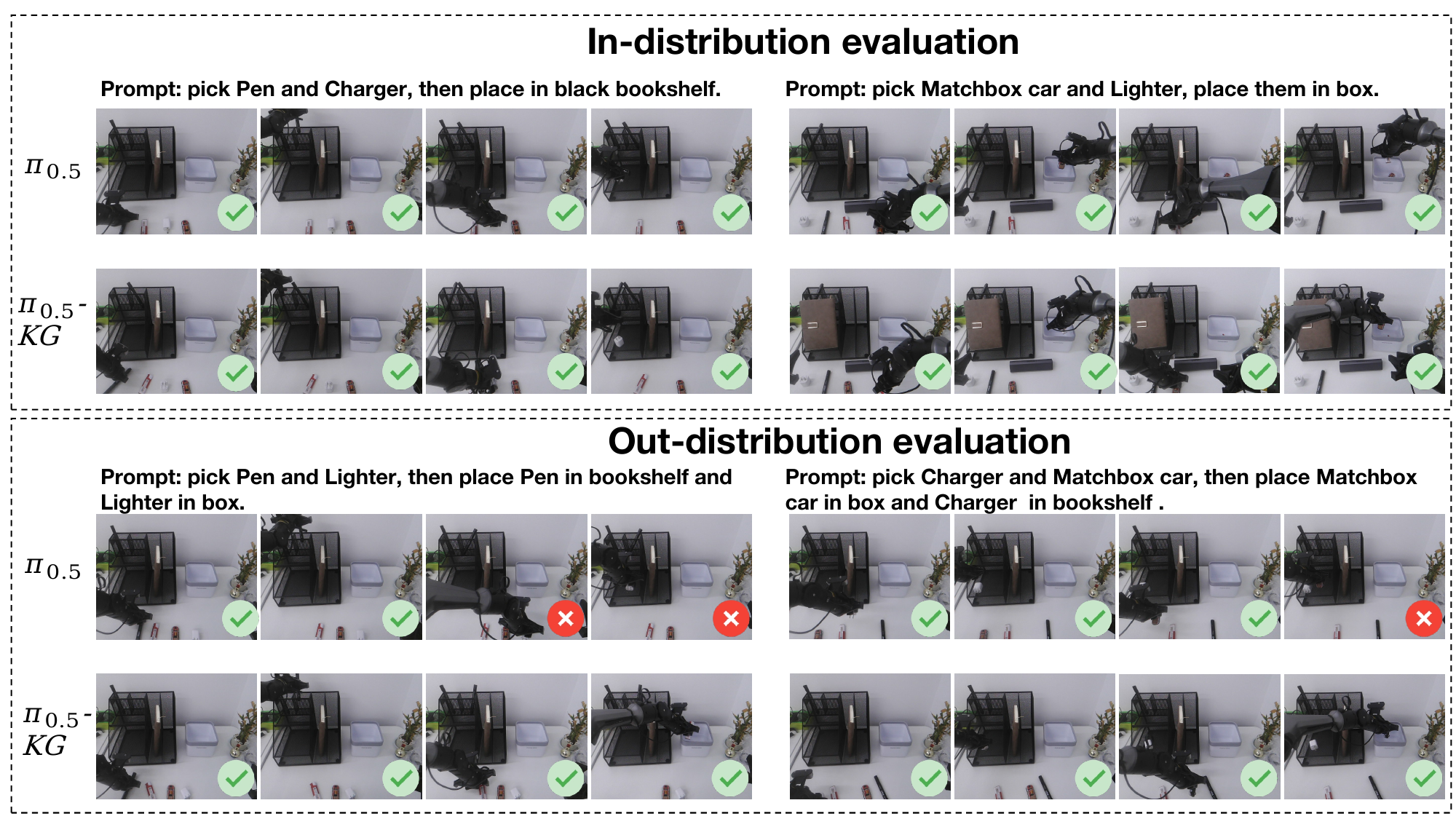}
    \caption{Qualitative examples from real-world expansion tasks. Both $\pi_{0.5}$ and $\pi_{0.5}$-KG succeed on in-distribution object pairs, while $\pi_{0.5}$ fails on unseen object-pair recombinations. In contrast, $\pi_{0.5}$-KG correctly follows the expanded instructions by grounding compositional object relations with retrieved linguistic priors.}
    \label{fig:real-expansion}
\end{figure*}

\paragraph{Expansion Tasks.}
Table~\ref{tab:real-expansion} reports the real-world expansion results. KG has little effect on in-distribution tasks where the base policies are already strong, but substantially improves unseen object-pair recombinations. It raises the average score from $72.4$ to $81.6$ for $\pi_{0}$ and from $80.0$ to $86.8$ for $\pi_{0.5}$, with the largest gains on compositional transfer tasks. Fig.~\ref{fig:real-expansion} shows the same trend qualitatively: $\pi_{0.5}$ succeeds on familiar pairings but fails after object-target relations are recombined, whereas $\pi_{0.5}$-KG completes the expanded instructions. This indicates that Key-Gram mainly benefits real-world generalization through object-concept recombination.

\begin{table*}[t]
  \caption{Real-world expansion-task results (\%). Gains in parentheses for KG variants are relative improvements over their corresponding base backbones. Task 1--5 correspond to two in-distribution settings, two unseen compositional pairings, and one sequential-adaptation setting.}
  \label{tab:real-expansion}
  \centering
  \small
  \setlength{\tabcolsep}{5pt}
  \begin{tabular}{lcccccc}
    \toprule
    Model & Task 1 & Task 2 & Task 3 & Task 4 & Task 5 & Avg. \\
    \midrule
    $\pi_{0}$
      & 92.0 & 90.0 & 52.0 & 48.0 & 80.0 & 72.4 \\
    $\pi_{0.5}$
      & 92.0 & 92.0 & 64.0 & 66.0 & 86.0 & 80.0 \\
    \midrule
    $\pi_{0}$-KG (ours)
      & 90.0 {\scriptsize(-2.2\%)}
      & 92.0 {\scriptsize(+2.2\%)}
      & 70.0 {\scriptsize(+34.6\%)}
      & 68.0 {\scriptsize(+41.7\%)}
      & 88.0 {\scriptsize(+10.0\%)}
      & 81.6 {\scriptsize(+12.7\%)} \\
    $\pi_{0.5}$-KG (ours)
      & 96.0 {\scriptsize(+4.3\%)}
      & 92.0 {\scriptsize(+0.0\%)}
      & 76.0 {\scriptsize(+18.8\%)}
      & 80.0 {\scriptsize(+21.2\%)}
      & 90.0 {\scriptsize(+4.7\%)}
      & 86.8 {\scriptsize(+8.5\%)} \\
    \bottomrule
  \end{tabular}
\end{table*}

\subsection{Ablation Study}
\label{sec:exp-ablation}

We conduct a layer-placement ablation on RoboTwin2.0 with the $\pi_{0.5}$ backbone to determine where Key-Gram should be injected. The ablation uses four representative tasks, \emph{hanging mug}, \emph{move stapler pad}, \emph{pick dual bottles}, and \emph{stack blocks three}, covering hanging, pushing, dual-object selection, and multi-object stacking. The final score is computed by weighting Easy and Hard settings with a ratio of $1{:}9$.

\begin{figure*}[t]
    \centering
    \includegraphics[width=\linewidth]{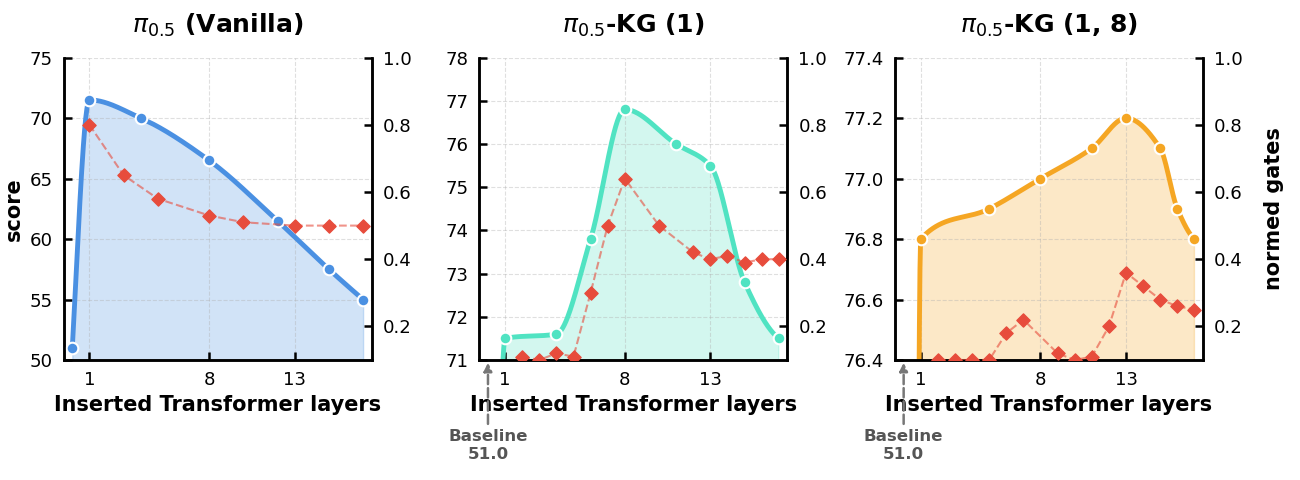}
    \caption{Layer-placement ablation on RoboTwin2.0. Shaded curves denote the weighted task score, while diamond markers denote the normalized gate activation of the probed Key-Gram module. The left and right axes correspond to score and normalized gates, respectively.}
    \label{fig:keygram-layer-ablation}
\end{figure*}

As shown in Table~\ref{tab:layer-ablation} and Fig.~\ref{fig:keygram-layer-ablation}, early insertion is most critical. A single insertion at Layer~1 improves the weighted score from $51.0$ to $71.5$, indicating that language-derived memory is most useful before the backbone spends substantial depth on instruction-related feature construction. Greedily adding Layer~8 further improves the score to $76.8$, while adding Layer~13 yields the best score of $77.2$. The marginal gain from $(1,8)$ to $(1,8,13)$ suggests diminishing returns from deeper memory injection.

Gate probing in Fig.~\ref{fig:keygram-layer-ablation} supports this trend. In the vanilla layer sweep, performance peaks at shallow layers and decreases for deeper insertions, although normalized gates remain above a small floor around $0.1$, indicating that deeper memories are still active but less influential. After one layer is selected, gates in nearby subsequent layers are strongly suppressed in the next probing stage, suggesting that the inserted module has already absorbed much of the useful linguistic prior for downstream blocks. This explains why the greedy search selects increasingly separated layers and finally adopts the $(1,8,13)$ configuration.

\section{Conclusion}
\label{sec:conclusion}

We presented Key-Gram, an external conditional memory framework for language-conditioned robot manipulation. Key-Gram separates instruction-side world-knowledge retrieval from scene-side visual reasoning by decomposing instructions into task-specific key-grams, retrieving static linguistic priors through deterministic hashed lookup, and injecting them into selected Transformer layers. This design reduces dense token-level modality competition and allows the visual backbone to focus on future-state reasoning and control.

Experiments on RoboTwin2.0, LIBERO, LIBERO-Plus, and real-world dual-arm tasks show that Key-Gram consistently improves both $\pi_{0}$ and $\pi_{0.5}$, with larger gains in long-horizon manipulation, distribution-expanded settings, and compositional object-pair transfer. Ablations further show that early memory injection is most effective, supporting the role of external linguistic priors in reducing instruction-related burden on the backbone. Overall, Key-Gram suggests a lightweight and extensible direction for organizing world knowledge in embodied control.

\begin{table*}[t]
  \caption{Layer-placement ablation results on RoboTwin2.0(\%). Numbers in parentheses denote the Transformer layers where Key-Gram is inserted. The score is computed as the average over the four tasks, with Easy and Hard settings weighted by a ratio of 1:9.}
  \label{tab:layer-ablation}
  \centering
  \small
  \setlength{\tabcolsep}{4pt}
  \begin{tabular}{lccccccccc}
    \toprule
    & \multicolumn{2}{c}{Hanging Mug}
    & \multicolumn{2}{c}{Move Stapler Pad}
    & \multicolumn{2}{c}{Pick Dual Bottles}
    & \multicolumn{2}{c}{Stack Blocks Three}
    &  \\
    \cmidrule(r){2-3}
    \cmidrule(r){4-5}
    \cmidrule(r){6-7}
    \cmidrule(r){8-9}
    Model
      & Easy & Hard
      & Easy & Hard
      & Easy & Hard
      & Easy & Hard
      & Score \\
    \midrule
    $\pi_{0.5}$ (Vanilla)
      & 18 & 17
      & 56 & 42
      & 93 & 63
      & 91 & 76
      & 51.0 \\
    $\pi_{0.5}$-KG (1)
      & 30 & 32
      & 83 & 79
      & 97 & 86
      & 94 & 87
      & 71.5 \\
    $\pi_{0.5}$-KG (1, 8)
      & 35 & 36
      & 88 & 85
      & 99 & 95
      & 96 & 90
      & 76.8 \\
    $\pi_{0.5}$-KG (1, 8, 13)
      & 38 & 35
      & 90 & 85
      & 98 & 98
      & 89 & 90
      & 77.2 \\
    \bottomrule
  \end{tabular}
\end{table*}

\newpage
\appendix

\section{Technical appendices and supplementary material}

\subsection{Full RoboTwin2.0 Results}
\label{app:robotwin-details}

\begingroup
\small
\setlength{\tabcolsep}{3pt}
\renewcommand{\arraystretch}{1.05}
\begin{longtable}{lcccccccc}
  \caption{Full RoboTwin2.0 results (\%). Gains in parentheses for KG variants are relative improvements over their corresponding base backbones.}
  \label{tab:robotwin-full-results}\\
  \toprule
  & \multicolumn{2}{c}{$\pi_{0}$}
  & \multicolumn{2}{c}{$\pi_{0.5}$}
  & \multicolumn{2}{c}{$\pi_{0}$-KG (ours)}
  & \multicolumn{2}{c}{$\pi_{0.5}$-KG (ours)} \\
  \cmidrule(r){2-3}
  \cmidrule(r){4-5}
  \cmidrule(r){6-7}
  \cmidrule(r){8-9}
  Task & Easy & Hard & Easy & Hard & Easy & Hard & Easy & Hard \\
  \midrule
  \endfirsthead

  \toprule
  & \multicolumn{2}{c}{$\pi_{0}$}
  & \multicolumn{2}{c}{$\pi_{0.5}$}
  & \multicolumn{2}{c}{$\pi_{0}$-KG (ours)}
  & \multicolumn{2}{c}{$\pi_{0.5}$-KG (ours)} \\
  \cmidrule(r){2-3}
  \cmidrule(r){4-5}
  \cmidrule(r){6-7}
  \cmidrule(r){8-9}
  Task & Easy & Hard & Easy & Hard & Easy & Hard & Easy & Hard \\
  \midrule
  \endhead

    adjust\_bottle & 99 & 95 & 100 & 99 & 99 & 98 & 100 & 99 \\
    beat\_block\_hammer & 79 & 84 & 96 & 93 & 92 & 90 & 100 & 94 \\
    blocks\_ranking\_rgb & 80 & 63 & 92 & 85 & 87 & 81 & 94 & 89 \\
    blocks\_ranking\_size & 14 & 5 & 49 & 26 & 55 & 36 & 71 & 60 \\
    click\_alarmclock & 77 & 68 & 98 & 89 & 75 & 78 & 99 & 92 \\
    click\_bell & 71 & 48 & 99 & 66 & 88 & 80 & 98 & 87 \\
    dump\_bin\_bigbin & 88 & 83 & 92 & 97 & 90 & 85 & 94 & 89 \\
    grab\_roller & 98 & 94 & 100 & 100 & 100 & 98 & 100 & 98 \\
    handover\_block & 47 & 31 & 66 & 57 & 70 & 58 & 75 & 61 \\
    handover\_mic & 97 & 97 & 98 & 97 & 97 & 90 & 98 & 96 \\
    hanging\_mug & 14 & 11 & 18 & 17 & 31 & 29 & 38 & 36 \\
    lift\_pot & 80 & 72 & 96 & 85 & 92 & 91 & 95 & 97 \\
    move\_can\_pot & 68 & 48 & 51 & 55 & 71 & 67 & 88 & 74 \\
    move\_pillbottle\_pad & 67 & 46 & 84 & 61 & 75 & 63 & 96 & 81 \\
    move\_playingcard\_away & 74 & 65 & 96 & 84 & 88 & 80 & 94 & 86 \\
    move\_stapler\_pad & 41 & 24 & 56 & 42 & 73 & 67 & 91 & 83 \\
    open\_laptop & 71 & 81 & 90 & 96 & 90 & 93 & 92 & 96 \\
    open\_microwave & 4 & 32 & 34 & 77 & 69 & 78 & 80 & 84 \\
    pick\_diverse\_bottles & 69 & 31 & 81 & 71 & 85 & 66 & 92 & 79 \\
    pick\_dual\_bottles & 59 & 37 & 93 & 63 & 91 & 83 & 100 & 97 \\
    place\_a2b\_left & 43 & 47 & 87 & 82 & 79 & 64 & 85 & 85 \\
    place\_a2b\_right & 39 & 34 & 87 & 84 & 74 & 57 & 86 & 87 \\
    place\_bread\_basket & 62 & 46 & 77 & 64 & 70 & 65 & 83 & 79 \\
    place\_bread\_skillet & 66 & 49 & 85 & 66 & 76 & 72 & 80 & 77 \\
    place\_burger\_fries & 81 & 76 & 94 & 87 & 94 & 81 & 92 & 88 \\
    place\_can\_basket & 55 & 46 & 62 & 62 & 86 & 73 & 92 & 77 \\
    place\_cans\_plasticbox & 63 & 45 & 94 & 84 & 72 & 63 & 94 & 85 \\
    place\_container\_plate & 97 & 92 & 99 & 95 & 95 & 94 & 96 & 95 \\
    place\_dual\_shoes & 59 & 51 & 75 & 75 & 78 & 79 & 86 & 83 \\
    place\_empty\_cup & 91 & 85 & 100 & 99 & 90 & 94 & 96 & 98 \\
    place\_fan & 66 & 71 & 87 & 85 & 78 & 86 & 93 & 88 \\
    place\_mouse\_pad & 20 & 20 & 60 & 39 & 52 & 41 & 75 & 70 \\
    place\_object\_basket & 67 & 70 & 80 & 76 & 77 & 70 & 80 & 78 \\
    place\_object\_scale & 57 & 52 & 86 & 80 & 73 & 68 & 87 & 82 \\
    place\_object\_stand & 82 & 68 & 91 & 85 & 88 & 83 & 91 & 87 \\
    place\_phone\_stand & 49 & 53 & 81 & 81 & 77 & 72 & 84 & 83 \\
    place\_shoe & 76 & 76 & 92 & 93 & 89 & 91 & 99 & 95 \\
    press\_stapler & 44 & 37 & 87 & 83 & 64 & 75 & 90 & 85 \\
    put\_bottles\_dustbin & 65 & 56 & 84 & 79 & 82 & 76 & 87 & 85 \\
    put\_object\_cabinet & 73 & 60 & 80 & 79 & 82 & 75 & 89 & 80 \\
    rotate\_qrcode & 74 & 70 & 89 & 87 & 88 & 83 & 86 & 84 \\
    scan\_object & 55 & 42 & 72 & 65 & 71 & 66 & 80 & 70 \\
    shake\_bottle\_horizontally & 98 & 92 & 99 & 99 & 92 & 93 & 97 & 98 \\
    shake\_bottle & 94 & 91 & 99 & 97 & 95 & 92 & 99 & 96 \\
    stack\_blocks\_three & 72 & 52 & 91 & 76 & 91 & 83 & 95 & 91 \\
    stack\_blocks\_two & 93 & 79 & 97 & 100 & 98 & 91 & 100 & 99 \\
    stack\_bowls\_three & 77 & 75 & 77 & 71 & 82 & 75 & 87 & 79 \\
    stack\_bowls\_two & 94 & 95 & 95 & 96 & 99 & 95 & 98 & 98 \\
    stamp\_seal & 46 & 33 & 79 & 55 & 67 & 70 & 89 & 83 \\
    turn\_switch & 41 & 42 & 62 & 54 & 40 & 42 & 60 & 56 \\
  \midrule
    Avg. $_{\mathrm{H}=1}$
      & 66.5 & 61.6
      & 85.1 & 80.2
      & 80.4 {\scriptsize(+20.9\%)} & 77.9 {\scriptsize(+26.4\%)}
      & 90.8 {\scriptsize(+6.6\%)} & 87.0 {\scriptsize(+8.5\%)} \\
    Avg. $_{\mathrm{H}=2}$
      & 66.1 & 54.7
      & 79.3 & 73.0
      & 80.5 {\scriptsize(+21.7\%)} & 72.7 {\scriptsize(+33.0\%)}
      & 86.3 {\scriptsize(+8.7\%)} & 80.3 {\scriptsize(+10.0\%)} \\
    Avg. $_{\mathrm{H}=3}$
      & 61.6 & 50.2
      & 78.6 & 67.4
      & 79.4 {\scriptsize(+28.9\%)} & 70.2 {\scriptsize(+39.8\%)}
      & 86.8 {\scriptsize(+10.4\%)} & 80.8 {\scriptsize(+19.9\%)} \\
    Avg. $_{\mathrm{50\ Tasks}}$
      & 65.9 & 58.4
      & 82.7 & 76.8
      & 80.3 {\scriptsize(+21.9\%)} & 75.6 {\scriptsize(+29.5\%)}
      & 89.0 {\scriptsize(+7.6\%)} & 84.4 {\scriptsize(+9.9\%)} \\
  \bottomrule
\end{longtable}
\endgroup

\subsection{Key-Gram Extraction Prompt}
\label{app:keygram-prompt}

For the RoboTwin2.0 dataset, we use an external language-model parser to convert each natural-language manipulation instruction into a fixed-size set of key-grams. Since RoboTwin2.0 contains diverse object-centric and long-horizon manipulation commands, the prompt is designed to encourage short, compositional, and action-centered phrases that preserve task-relevant semantic roles, such as actions, objects, attributes, sources, targets, and spatial relations. Unless otherwise specified, we extract exactly eight key-grams for each RoboTwin2.0 instruction. The complete prompt used in our experiments is shown below.

\begin{verbatim}
SYSTEM_PROMPT = """
You are an expert parser for embodied manipulation instructions.

Your task is to convert one instruction into exactly 8 composed keywords.

These keywords must be short, natural, semantically meaningful phrases for embodied actions, not fragmented n-grams.

Rules:
1. Output exactly 8 keywords.
2. Each keyword must contain 2 to 4 words.
3. Prefer high-information phrases that combine multiple semantic roles in one phrase.
4. Prefer action-centered phrases over static descriptive phrases whenever possible.
5. At least 3 of the 8 keywords must explicitly contain an action verb.
6. Prefer these phrase types, in this priority order:
   a. verb + object + relation/target/source
   b. verb + particle + object
   c. verb + prep + object
   d. object + prep + object
   e. attribute + object
7. A good keyword should ideally compress 2 or more semantic elements, such as:
   - action + object
   - action + object + source
   - action + object + target
   - object + attribute
   - object + location
8. Use standalone static noun phrases only when they add important information that is not already covered elsewhere.
9. Use at most 5 standalone noun phrases.
10. If a static phrase can be replaced by a more informative action phrase, prefer the action phrase.
11. Prefer phrases like:
   - "pick up"
   - "pick bowl from drawer"
   - "pick up bowl"
   - "place bowl on plate"
   - "bowl in top drawer"
   - "black bowl"
12. Avoid:
   - fragmented phrases
   - fake combinations across unrelated spans
   - pronoun-centered phrases like "place it on"
   - low-information phrases
   - too many static environment phrases
   - duplicated semantics across multiple keywords
   - more than 4 words in a keyword
   - less or more than 8 keywords
13. Do not explain anything.
14. Return valid JSON only.

Example:
Instruction:
pick up the green sponge from the sink and wipe the wooden table near the window

Output:
{
  "keywords": [
    "pick and wipe",
    "pick sponge from sink",
    "pick up sponge",
    "green sponge",
    "wipe wooden table",
    "wipe table near window",
    "table near window",
    "wooden table"
  ]
}

MUST FOLLOW:
- Do NOT less or more than 8 keywords.
- Do NOT use more than 4 words in a keyword.
Must check the keywords strictly follow the rules above.

"""
\end{verbatim}

\subsection{Model Configuration Details}
\label{appendix:model-config}

\paragraph{Simulation experiments.}
For the simulation experiments on LIBERO, LIBERO-Plus, and RoboTwin2.0, we choose the Key-Gram configuration according to the instruction length and linguistic complexity of these benchmarks. Specifically, we set the number of key-gram slots to $S=8$, the number of hash heads per slot to $H=4$, and the embedding width of each hash head to $d_h=32$. This gives a retrieved memory width of $d_m=S H d_h=1024$ for each inserted layer. For each hash-head table, we use a row capacity of $V=8192$. This capacity provides sufficient address space for more than four thousand object- and relation-level entities, reducing the probability of hash collisions and avoiding frequent updates of semantically unrelated entities to the same memory row. The final fusion stage uses a lightweight 1D convolution. To encourage interaction among all retrieved key-gram slots, we set the convolutional span to the slot budget, i.e., $w=S=8$. Key-Gram is inserted into Transformer layers $(1,8,13)$ under zero-based layer indexing, while the original VLA backbone and action expert are kept unchanged.

\begin{table}[htbp]
\centering
\small
\caption{Key-Gram configuration for simulation experiments. The layer indices are zero-based.}
\label{tab:keygram-config-sim}
\begin{tabular}{l c}
\toprule
\textbf{Configuration item} & \textbf{Value} \\
\midrule
Benchmarks & LIBERO, LIBERO-Plus, RoboTwin2.0 \\
Backbones & $\pi_{0}$, $\pi_{0.5}$ \\
Key-gram slots $S$ & 8 \\
Hash heads per slot $H$ & 4 \\
Head embedding width $d_h$ & 32 \\
Retrieved memory width $d_m=S H d_h$ & 1024 \\
Rows per hash-head table $V$ & 8192 \\
Number of hash-head tables per layer $S \times H$ & 32 \\
Embedding parameters per inserted layer $S H V d_h$ & 8.39M \\
Convolution type & 1D convolution \\
Convolutional span $w$ & 8 \\
Inserted Transformer layers & $(1,8,13)$ \\
Layer indexing & zero-based \\
\bottomrule
\end{tabular}
\end{table}

\paragraph{Real-world experiments.}
For the real-world dual-arm experiments, we use a slightly different allocation because the real-world instructions are generally longer than those in simulation, while the number of visually instantiated assets and object-level semantic variants is smaller than in simulation benchmarks. We therefore increase the number of key-gram slots to $S=16$ to better cover long-horizon and extensible-task instructions, but reduce the number of hash heads per slot to $H=2$. The head embedding width remains $d_h=32$, resulting in the same retrieved memory width $d_m=S H d_h=1024$ as in simulation. For consistency, each hash-head table also uses $V=8192$ rows. This setting allocates more slots to instruction decomposition while keeping the total number of hash-head tables unchanged, which balances long-instruction coverage and memory efficiency. The fusion module again uses a lightweight 1D convolution, with the convolutional span set to the slot budget, i.e., $w=S=16$, so that information from all key-gram slots can be jointly mixed before residual injection. As in simulation, Key-Gram is inserted into Transformer layers $(1,8,13)$ under zero-based layer indexing, and the backbone and action expert are kept unchanged.

\begin{table}[htbp]
\centering
\small
\caption{Key-Gram configuration for real-world robot experiments. The layer indices are zero-based.}
\label{tab:keygram-config-real}
\begin{tabular}{l c}
\toprule
\textbf{Configuration item} & \textbf{Value} \\
\midrule
Tasks & long-horizon tasks and extensible tasks \\
Backbones & $\pi_{0}$, $\pi_{0.5}$ \\
Key-gram slots $S$ & 16 \\
Hash heads per slot $H$ & 2 \\
Head embedding width $d_h$ & 32 \\
Retrieved memory width $d_m=S H d_h$ & 1024 \\
Rows per hash-head table $V$ & 8192 \\
Number of hash-head tables per layer $S \times H$ & 32 \\
Embedding parameters per inserted layer $S H V d_h$ & 8.39M \\
Convolution type & 1D convolution \\
Convolutional span $w$ & 16 \\
Inserted Transformer layers & $(1,8,13)$ \\
Layer indexing & zero-based \\
\bottomrule
\end{tabular}
\end{table}

\subsection{Limitations and Future work}
\label{app:limitations}

This work has some limitations. First, due to the high cost of large-scale simulation and real-world robot evaluation, we instantiate Key-Gram only on two open-source VLA backbones, $\pi_{0}$ and $\pi_{0.5}$. Although the results are consistent across RoboTwin2.0, LIBERO, LIBERO-Plus, and real-world dual-arm tasks, they do not yet fully establish the generality of Key-Gram across a broader range of VLA architectures or world-model-based manipulation policies. Then, the current framework relies on an external parser to extract task-specific key-grams from language instructions. While this provides a controllable interface for memory retrieval, extraction quality may affect downstream performance, especially for ambiguous, long, or highly compositional instructions. Future work should evaluate Key-Gram on more diverse backbones, larger real-world task suites, and longer continual-learning settings, while exploring parser-free or jointly learned key-gram discovery.

\subsection{Computational Resources}
\label{app:compute-resources}

For model training, we utilize a compute node equipped with 8 NVIDIA H200 GPUs for full-parameter training. Operating with a per-GPU batch size of 16 (yielding a total global batch size of 128), the VRAM consumption is approximately 135 GB per GPU. Under this hardware configuration, the training throughput reaches approximately 128 samples per second. For the complete RoboTwin2.0 dataset, training typically requires 1 to 1.5 days per epoch.


\end{document}